\documentclass[conference]{IEEEtran}
\ifCLASSINFOpdf
\else
\fi
\usepackage{floatflt}
\usepackage{graphics,cite,times,mathptmx,amsmath,amssymb,bm, listings, graphicx}
\usepackage{algorithm}
\usepackage{algorithmic}
\usepackage{epsfig}
\usepackage{subfigure}

\begin{document}
%
\title{Modeling Based on Elman Wavelet Neural Network for Class-D Power Amplifiers}

\author{\IEEEauthorblockN{Li Wang$^{1}$, Jie Shao$^{1}$, Yaqin Zhong$^{1}$, Weisong Zhao$^{1}$, Reza Malekian$^{2}$}
\IEEEauthorblockA{$^1$College of Electronic and Information Engineering, Nanjing University of Aeronautics and Astronautics, Nanjing, 210016, China \\
$^2$Department of Electrical, Electronic and Computer Engineering, University of Pretoria, South Africa
\\
Email: reza.malekian@ieee.org}
}


%


\maketitle

\begin{abstract}
In Class-D Power Amplifiers (CDPAs), the power supply noise can intermodulate with the input signal, manifesting into power-supply induced intermodulation distortion (PS-IMD) and due to the memory effects of the system, there exist asymmetries in the PS-IMDs. In this paper, a new behavioral modeling based on the Elman Wavelet Neural Network (EWNN) is proposed to study the nonlinear distortion of the CDPAs. In EWNN model, the Morlet wavelet functions are employed as the activation function and there is a normalized operation in the hidden layer, the modification of the scale factor and translation factor in the wavelet functions are ignored to avoid the fluctuations of the error curves. When there are $30$ neurons in the hidden layer, to achieve the same square sum error (SSE) $\epsilon_{min}=10^{-3}$, EWNN needs $31$ iteration steps, while the basic Elman neural network (BENN) model needs 86 steps. The Volterra-Laguerre model has 605 parameters to be estimated but still can't achieve the same magnitude accuracy of EWNN. Simulation results show that the proposed approach of EWNN model has fewer parameters and higher accuracy than the Volterra-Laguerre model and its convergence rate is much faster than the BENN model.
\end{abstract}


\section{Introduction}

The Class-D Power Amplifiers (CDPAs) are increasingly ubiquitous largely because of their significantly higher power efficiency attribute compared to their linear counterparts ~\cite{1}. The PWM (Pulse Width Modulation) is the most prevalent modulation technique. The output transistors of CDPAs operate in the ohmic and cut-off regions, make the output voltage contains ripple ~\cite{2}, whose power spectral density is high at multiples of switching frequency. But noise in the power-supply has a greater impact than the switching frequency ~\cite{3}. One of the reasons is that the power supply noise may intermodulate with the input signal, manifesting into power-supply induced intermodulation distortion (PS-IMD) ~\cite{4,5}, and that, in some instances, the PS-IMD can be significantly larger than the output distortion component at supply noise frequency. As a drawback of CDPAs, it's necessary to have a research on the PS-IMD. A feasible way is to modeling the CDPAs' nonlinearity accurately and analysis the spectrum of model's output.

Behavioral modeling ~\cite{6} is often used in PA's nonlinear analysis because it provides a convenient and efficient mean to predict system-level performance without the computational complexity of full circuit simulation or physical level analysis, thereby significantly speeding up the analysis process. There has been intensive research in memoryless nonlinear behavioral modeling of PAs, however, memory effects in real PAs often arise due to thermal effects and large time constants in dc-bias circuits ~\cite{7}. In the simulations, the PS-IMD is asymmetrical between the upper and lower sidebands obviously, which should be caused by the memory effects. Based on this, behavioral models which have memory effects are used, such as Volterra series expansion models ~\cite{8} and the neural network models ~\cite{9,18}.

Anding Zhu and Thomas J. Brazil proposed a behavioral model for power amplifiers in ~\cite{10}, by projecting the classical Volterra series onto a set of Orthogonal Basis Functions, namely, the Laguerre functions. This approach enables a substantial reduction in the number of parameters involved, and allows the reproduction of both transient and steady-state behavior of power amplifiers ~\cite{16} with excellent accuracy.

The basic Elman neural network (BENN) ~\cite{11} is a partial recurrent network model first proposed by Elman in 1990. Its back-forward loop employs context layer which is sensitive to the history of input data, so the network can manifest the memory effect of the power amplifiers. Since signals of interest can usually be expressed using wavelet decompositions ~\cite{12}, and signal processing algorithms can be performed by adjusting only the corresponding wavelet coefficients, we propose a behavioral modeling based on BENN and the wavelets ~\cite{13} in this paper, namely the Elman Wavelet Neural Network (EWNN) model ~\cite{14}. In EWNN model, the nonlinear Morlet wavelet functions are used as a substitute for the activation function of hidden layer neurons in the BENN model. The input data before wavelet transformation is normalized to guarantee the convergence of the algorithm. In the learning process of the EWNN model, the update of the scale factor and translation factor in the wavelets are ignored as they do little contribution to the convergence of the algorithm. The update of two parameters causes a lot of fluctuations to the square sum error (SSE) and may lead the SSE to a local minimum. Combining with the fast convergence of wavelet networks, the proposed modeling is more effective for power amplifiers than the BENN model.

The remainder of this paper is organized as follows. The asymmetry of PS-IMD is analyzed in Sect. 2. The theory of the BENN model is illustrated in Sect. 3. Then Sect. 4 gives the principle of the EWNN model. The simulation results and the discussion are given in Sect. 5.

\section{THE ASYMMETRY OF PS-IMD IN THE HALF-BRIDGE CDPA}
In this paper, the half-bridge CDPA circuit showed in Fig. 1 ~\cite{2} is used to analyze the asymmetry of the PS-IMD.
\begin{figure}
\centering
\includegraphics[width=8cm, height=5cm]{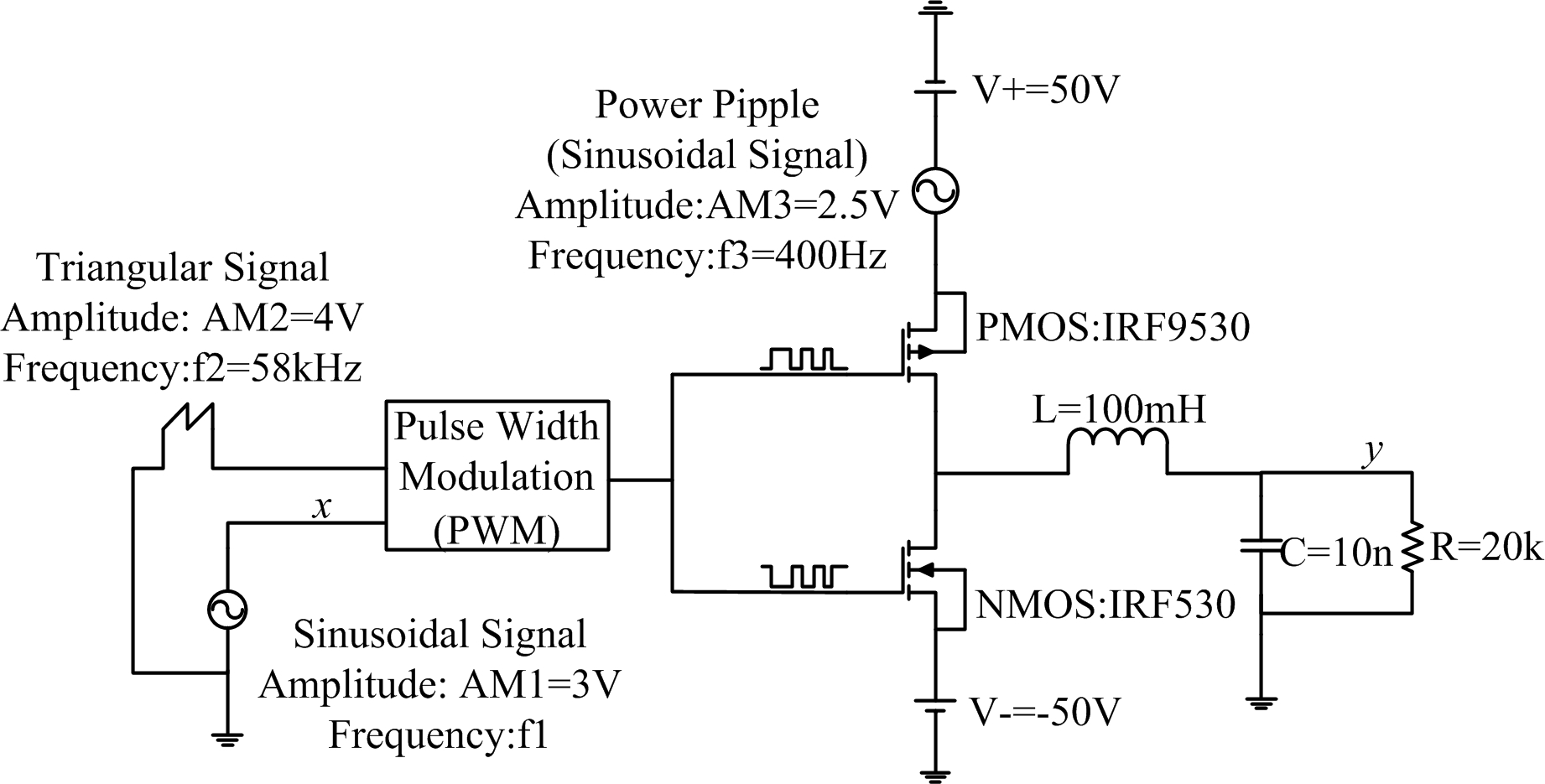}
\caption{Structure of the half-bridge D-class power amplifier}
\label{1}
\end{figure}

The amplitude of input sinusoidal signal is $3V$. The triangular signal has the frequency of 58kHz and the amplitude of $4V$. The power supply is added with a $5\%$ sinusoidal voltage ripple with the frequency of $400Hz$. The cut-off frequency of the LC low-pass filter is about $10kHz$. Since the voltage ripple can't be filtered by the LC low-pass filter, there exist PS-IMDs in the output signal.

In two-tone or multi-tone input PAs, there exist asymmetries in lower and upper sidebands and the intermodulation distortion magnitude variation depending on input frequency interval. It is known that these phenomena come from the memory effects ~\cite{7}, which means that the output depends not only on the input signal at the moment but also on the history of past input levels. Adding a sinusoidal power ripple to the power supply of the open-loop CDPA circuit, the expression of the PWM signal before the low-pass filter is given in ~\cite{4}. It shows that the PS-IMD should be symmetrical in the output signal. But in actual, there exists obvious asymmetries in lower and upper PS-IMDs. We attribute the asymmetry to the memory effects according to the system of two-tone input PAs. Fig. 2 gives the output spectrum of the circuit when the input frequency is $3700$Hz.
\begin{figure}
\centering
\includegraphics[width=8cm, height=7cm]{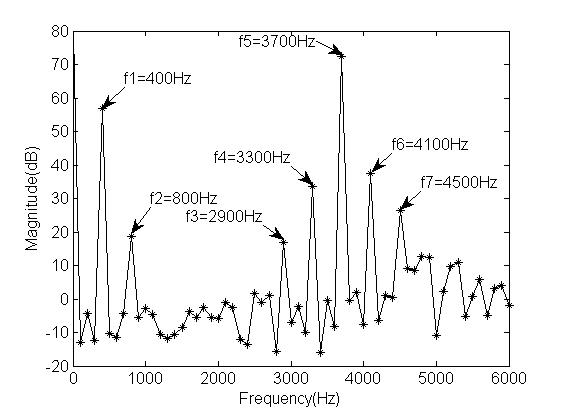}
\caption{Output spectrum with the input signal's frequency of $3700Hz$}
\label{2}
\end{figure}
As marked in Fig. 2, $f_{1}=400Hz$ is the frequency of power ripple, $f_{2}=800Hz$ is the second harmonic component of $f_{1}.f_{5}=3700Hz$ is the frequency of input sinusoidal signal.

$f_{3}=f_{5}-2f_{1}$ and $f_{7}=f_{5}+2f_{1}$ are the third-order intermodulation distortion (PS-IMD3), $f_{4}=f_{5}-f_{1}$ and $f_{6}=f_{5}+f_{1}$ are the second-order intermodulation distortion (PS-IMD2). It can be seen that there exists asymmetry in PS-IMD2 and PS-IMD3.

Fig. 3 gives the measured asymmetry results in PS-IMD2 and PS-IMD3 by sweeping the input signal's frequency from $1.9kHz$ to $4.3kHz$.
\begin{figure}
\centering
\includegraphics[width=8cm, height=7cm]{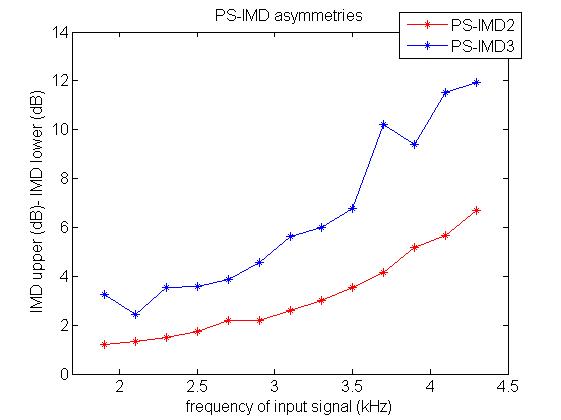}
\caption{PS-IMD2 and PS-IMD3 asymmetries}
\label{3}
\end{figure}
It can be seen from Fig.3 that the measured PS-IMD2 difference between lower and upper terms ranges from $1.21dB$ to $6.7dB$. The measured PS-IMD3 difference between lower and upper terms ranges from $2.439dB$ to $11.92dB$. The amount of the asymmetry depends on the frequency spacing between the input signal and the power ripple. With the increasing of the frequency spacing, the asymmetry grows.

\section{THE BASIC ELMAN NEURAL NETWORK}
Considering the multi-input and multi-output system, the BENN model is illustrated in Fig. 4.
\begin{figure}
\centering
\includegraphics[width=8cm, height=6cm]{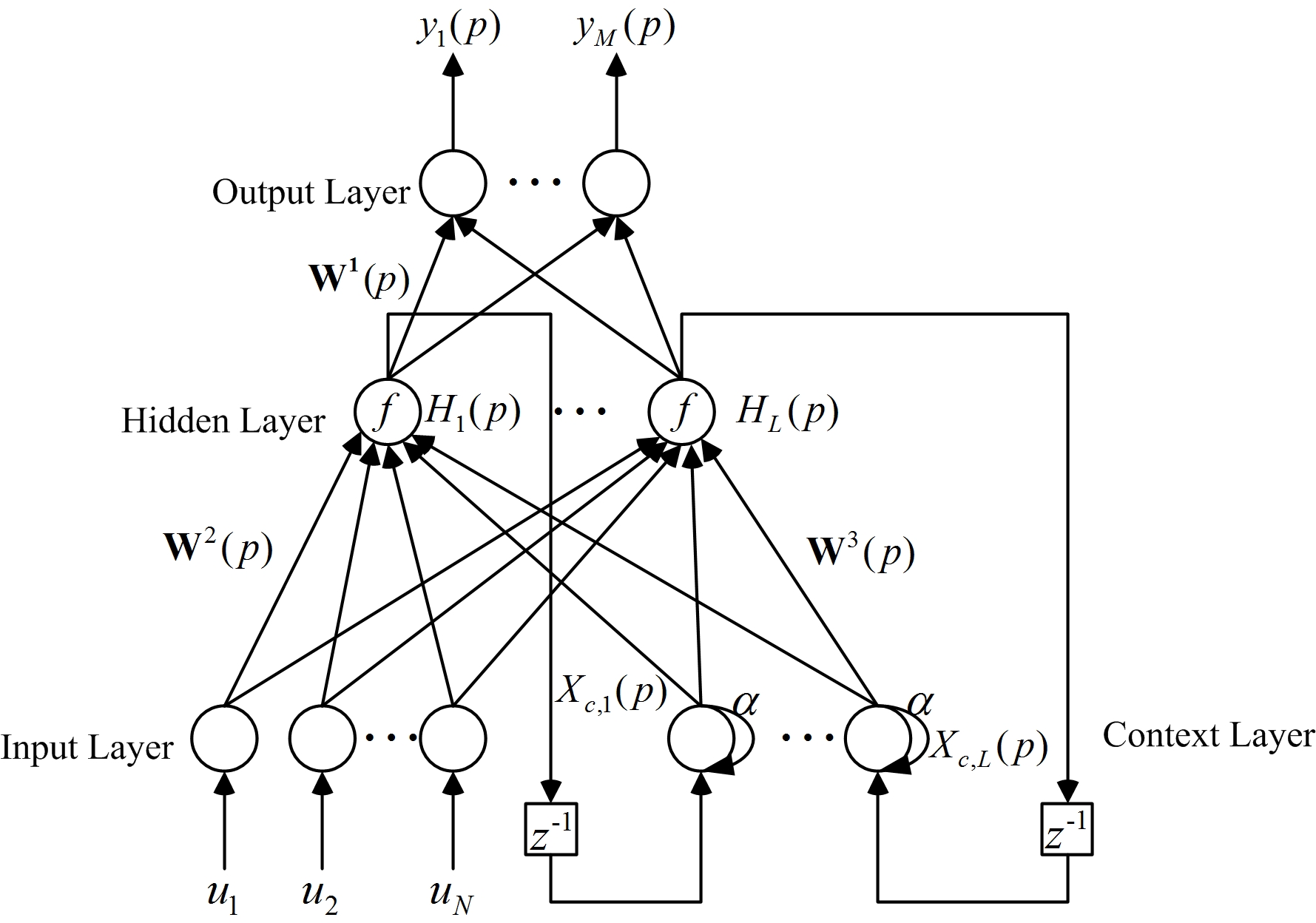}
\caption{Structure of the BENN model}
\label{4}
\end{figure}
The BENN model is composed of four layers: input layer, hidden layer, context layer, and output layer. The input layer has N input nodes. It accepts the input variables and transmits to the hidden layer. The hidden layer has L nodes and contains the transfer function f. The context layer is the feedback loop of hidden layer with a self-loop coefficient $\alpha$ and it has L neural nodes, too. The output of the context layer at $p-th$ learning step is related to the output of the hidden layer at $(p-1)th$ step. The output layer has M nodes and the output $y_{j} \quad (j=1,2,..., M)$ is the linear combination of the output of the hidden layer. There are three kind of weight in the network: $W^1$ is the $L \times M$ dimensional weight matrix from the hidden layer to the output layer. $W^2$ is the $N \times L$ dimensional weight matrix from the input layer to the hidden layer. $W^3$ is the $L \times L$ dimensional weight matrix from the context layer to the hidden layer. The dynamical equations [9] of the BENN model are as follows:

\begin{equation}
y(p)=W^1(p)H(p)
\end{equation}

\begin{equation}
H(p)=f[W^2(p)u+W^3(p)X_{c}(p)]
\end{equation}

\begin{equation}
X_{c}(p)=\alpha H(p-1)
\end{equation}
Where $p$ is the number of iteration steps and $f(x)$  usually represents the Sigmoid function.
\begin{equation}
f(x)=\frac{1}{1+e^{-x}}
\end{equation}
By using the gradient descent (GD) method [15], the weight values are adjusted so that the SSE is minimized after training cycles. Suppose that the p-th iteration output of the network is $y(p)$ , the objective performance error-function is defined as
\begin{equation}
E(p)=\frac{1}{2}[(y_{d}-y(p))^T(y_{d}-y(p))]
\end{equation}
Where $y_{d}$  is the desired output of the model. The partial derivative of error-function with respect to the weight parameters are as follows
\begin{equation}
\Delta{W_{im}}^1(p)=-\eta_{1}\frac{\partial E(p)}{\partial W_{im}^1(p)}=\eta_{1}\delta_{m}^o(p)H_{i}(p)
\end{equation}
\begin{equation}
\Delta W_{ji}^2(p)=-\eta_{2} \frac{\partial E(p)}{\partial W_{ji}^2(p)}=\eta_{2} \delta_{i}^{h}(p)\frac {\partial H_{i}(p)}{\partial W_{ji}^{2}(p)}
\end{equation}
\begin{equation}
\Delta W_{ki}^3(p)=-\eta_{3} \frac{\partial E(p)}{\partial W_{ki}^3(p)}=\eta_{3} \delta_{i}^{h}(p)\frac {\partial H_{i}(p)}{\partial W_{ki}^{3}(p)}
\end{equation}
With
\begin{equation}
\delta_{m}^o(p)=y_{d,m}-y_{m}(p)
\end{equation}
\begin{equation}
\delta_{i}^{h}(p)=\sum_{m=1}^{M} (\delta_{m}^o(p)W_{im}^{1}(p))
\end{equation}
\begin{equation}
\frac{\partial H_{i}(p)}{\partial W_{ji}^2(p)}=f_{i}^{'}(.)[u_{j}+\alpha .W_{ii}^{3}(p).\frac{\partial H_{i}(p-1)}{\partial W_{ij}^2(p-1)}
\end{equation}
\small
\begin{equation}
\frac{\partial H_{i}(p)}{\partial W_{ki}^3(p)}=f_{i}^{'}(.)[\alpha .H_{k}(p-1)+\alpha .W_{ii}^3(p).\frac{\partial H_{i}(p-1)}{\partial W_{ki}^3(p-1)}
\end{equation}
Where $j$ represents the $j-th$ neuron of the input layer $(j=1,2,...,N)$; $i$ represents the $i-th$ neuron of the hidden layer $(i=1,2,...L)$; $k$ represents the $k-th$ neuron of the context layer $(k=1,2,...,L)$; $m$ represents the $m-th$ neuron of the output layer. $\eta_{1},\eta_{2},\eta_{3}$ represent the learning rate of $W^1,W^2,W^3$ respectively. $f_{i}^{'}$ is the derived function of the transfer function $f$.	

\section{THE ELMAN WAVELET NEURAL NETWORK}
\subsection{The structure of the Elman Wavelet Neural Network}
Since signals of interest can usually be expressed using wavelet decompositions, and signal processing algorithms can be performed by adjusting only the corresponding wavelet coefficients, we use the nonlinear wavelets as the substitute of the Sigmoid function in the hidden layer and propose a new model-the Elman Wavelet Neural Network (EWNN) model. The structure of EWNN is similar to BENN, the only difference is that the transfer function in the hidden layer is replaced by wavelet functions. EWNN has combined the properties such as attractor dynamics of RNN (Recurrent Neural Network) and the fast convergence of WNN (Wavelet Neural Network), it can capture the past information of the network and can adapt rapidly to sudden changes.
In this paper, the Morlet wavelet which is the Gauss wavelet of cosine modulation is chosen as the mother wavelet in the hidden layer. The mother wavelet $\Psi(x)$ and the wavelet transform $\Psi_{a_{i},b_{i}}z_{i}$ are defined as follows:
\begin{equation}
\Psi(x)=cos(1.75x).exp(-\frac {x^2}{2})
\end{equation}
\begin{equation}
\Psi_{{a_i},{b_i}}(z_{i})=cos(1.75z_{i}).exp(-\frac {z_{i}^2}{2})
\end{equation}
In this model, a very important step is to normalize $z(p)$ of the hidden layer. This operation ensures the convergence of the algorithm. If don't do this, the SSE will keep a large value instead of decreasing with the increasing of the iteration steps. The normalization function is
\begin{equation}
z_{i}(p)=\frac{z_{i}^{'}(p)}{max(|z^{'}(p)|)}
\end{equation}
With $z_{i}^{'}=(h_{i}-b_{i}/a_{i})$, where $h_{i}(p)=W_{ji}^2(p).u_{j}(p)+ W_{ki}^3(p)X_{c,k}(p)$ is the input of $i-th$ node in the hidden layer, and $a_{i},b_{i}$ , are termed as the scale factor and translation factor of the wavelets in the hidden layer, respectively.
The dynamical equation (2) for hidden layer is replaced by ~\cite{15}
\begin{equation}
H(p)=\Psi_{{a},{b}}[W^2(p)u+W^3(p)X_{c}(p)]
\end{equation}
Equations (11) and (12) are modified to
\begin{equation}
\frac{\partial H_{i}(p)}{\partial W_{ji}^2(p)}=\frac {\Psi^{'}_{{a_i},{b_i}}(z_{i})}{a_{i}(p)}[u_{j}+\alpha .W_{ii}^{3}(p).\frac{\partial H_{i}(p-1)}{\partial W_{ij}^2(p-1)}]
\end{equation}
\small
{\footnotesize
\begin{equation}
\frac{\partial H_{i}(p)}{\partial W_{ki}^3(p)}=\frac {\Psi^{'}_{{a_i},{b_i}}(z_{i})}{a_{i}(p)}[\alpha . H_{k}(p-1)+\alpha .W_{ii}^3(p). \frac{\partial H_{i}(p-1)}{\partial W_{ki}^3(p-1)}]
\end{equation}}
{\scriptsize
\begin{equation}
\Psi_{{a_i},{b_i}}^{'}(z_{i})=-1.75sin(1.75z_{i}).exp(-\frac {z_{i}^2}{2})-z_{i}cos(1.75z_{i}).exp(-\frac {z_{i}^2}{2})
\end{equation}}
The partial derivative of error-function with respect to $a$ and $b$ are
\begin{equation}
\Delta a_{i}(p)=-\eta_{4} \frac{\partial E(p)}{\partial a_{i}(p)}=\eta_{4} \delta_{i}^{h}(p)\frac{\partial H_{i}(p)}{\partial a_{i}(p)}
\end{equation}
\begin{equation}
\Delta b_{i}(p)=-\eta_{4} \frac{\partial E(p)}{\partial a_{i}(p)}=\eta_{4} \delta_{i}^{h}(p)\frac{\partial H_{i}(p)}{\partial b_{i}(p)}
\end{equation}
\begin{equation}
\frac{\partial H_{i}(p)}{\partial a_{i}(p)}=\Psi_{{a_i},{b_i}}^{'}(z_{i}).[-\frac{z_{i}}{a_{i}(p)}+\alpha.W_{ii}^{3}(p).\frac{\partial H_{i}(p-1)}{\partial a_{i}(p-1)}]
\end{equation}
\begin{equation}
\frac{\partial H_{i}(p)}{\partial a_{i}(p)}=\Psi_{{a_i},{b_i}}^{'}(z_{i}).[-\frac{1}{a_{i}(p)}+\alpha.W_{ii}^{3}(p).\frac{\partial H_{i}(p-1)}{\partial a_{i}(p-1)}]
\end{equation}
where $\eta_{4}$ and $\eta_{5}$ represent the learning rate of $a$ and $b$ respectively.
The initial value of the parameter vectors $a$ and $b$ is usually random. As the changes of $a$ and $b$  are unpredictable, the update of them will cause large fluctuations to SSE with the increase of iterative times. This may make the SSE curve fall into local minimum and stop the learning process at a wrong place. To avoid this, here, we take the parameters $a$ and $b$ as constants and ignore the modification of them. Which means when training the EWNN model, formulas (20) to (23) are ignored. The $a_{i}(p)$ in (17) and (18) are also set to 1 to avoid the fluctuations.
\subsection{The training of the EWNN model} By using the GD method updating the weight matrixes, the training steps to determine the optimal number of hidden neurons of the EWNN model are as follows:

Step1: Initialize the network. Choose a initial number for the neurons in hidden layer $L=10$. Set the weight matrix $W^1,W^2,W^3$ to zero matrixes and parameters $a$, $b$ to be randomly subject to the standard normal distribution. Set $\frac {\partial H_{i}(0)}{\partial W_{ji}^{2}(0)}=0$,$\frac {\partial H_{i}(0)}{\partial W_{ki}^{3}(0)}=0$. Determine the maximum number of iterations $N_{max}=100$ and the threshold value of SSE $\epsilon_{min}=10^{-3}$. The initial value of the context layer is $X_{c}(0)=0$. Set the self-loop coefficient of the context layer $\alpha=0.001$, the learning rate $\eta_{1}=\eta_{2}=\eta_{3}=0.01$

Step2: According to formula (1) (16) (3), calculate the output $y(p)$. Calculate the SSE of the $p-th$ iteration step $E(p)$, if $E(p)< \epsilon_{min}$ or the number of iteration $p\geq N_{max}$, end the training process, else execute step 3.

Step3: Acquire the adjustment values of the weight matrixes:$\Delta W_{im}^{1}(p)$,$\Delta W_{ji}^{2}(p)$, $\Delta W_{ki}^{3}(p)$. Then the parameters are updated as: $W_{im}^{1}(p+1)=W_{im}^{1}(p)+\Delta W_{im}^{1}(p)$, $W_{ji}^{2}(p+1)=W_{ji}^{2}(p)+\Delta W_{ji}^{2}(p)$, $W_{ki}^{3}(p+1)=W_{ki}^{3}(p)+\Delta W_{ki}^{3}(p)$. Jump to step 2.

Step4: With the weight matrix obtained in step3, calculate the final output $y$ of EWNN.

Step5: Change the number of hidden neurons and repeat step1 to step4. By continuously testing, find the most suitable number for the hidden neurons.
\section{SIMULATION RESULTS AND THE ANALYSIS}
The simulation data is acquired from the half-bridge CDPA circuit showed in Fig. 1. The frequency of the input signal is $3700Hz$. The data for modeling is achieved by sampling the circuit's input $x$ and output $y$ between $10ms$ and $20ms$ and the sampling frequency is $100kHz$. In both the BENN and EWNN model, the self-feedback coefficient $\alpha$ is set to $0.001$ and the learning rate $\eta_{1},\eta_{2}, \eta_{3}$ are set to $0.01$.

\subsection{Error curves of the BENN and the EWNN model}
 Firstly, a study on the relationship between the number of hidden neurons and the error curve of SSE with the increase of the iteration steps is done. Set a large number for iteration times $N_{max}=100$.

For the BENN model, make the number of hidden neurons $L$ increases from $10$ to $110$ with the interval of $10$. The error curves of SSE with the increase of the number of hidden neurons are showed in Fig. 5.
\begin{figure}
\centering
\includegraphics[width=9cm, height=6cm]{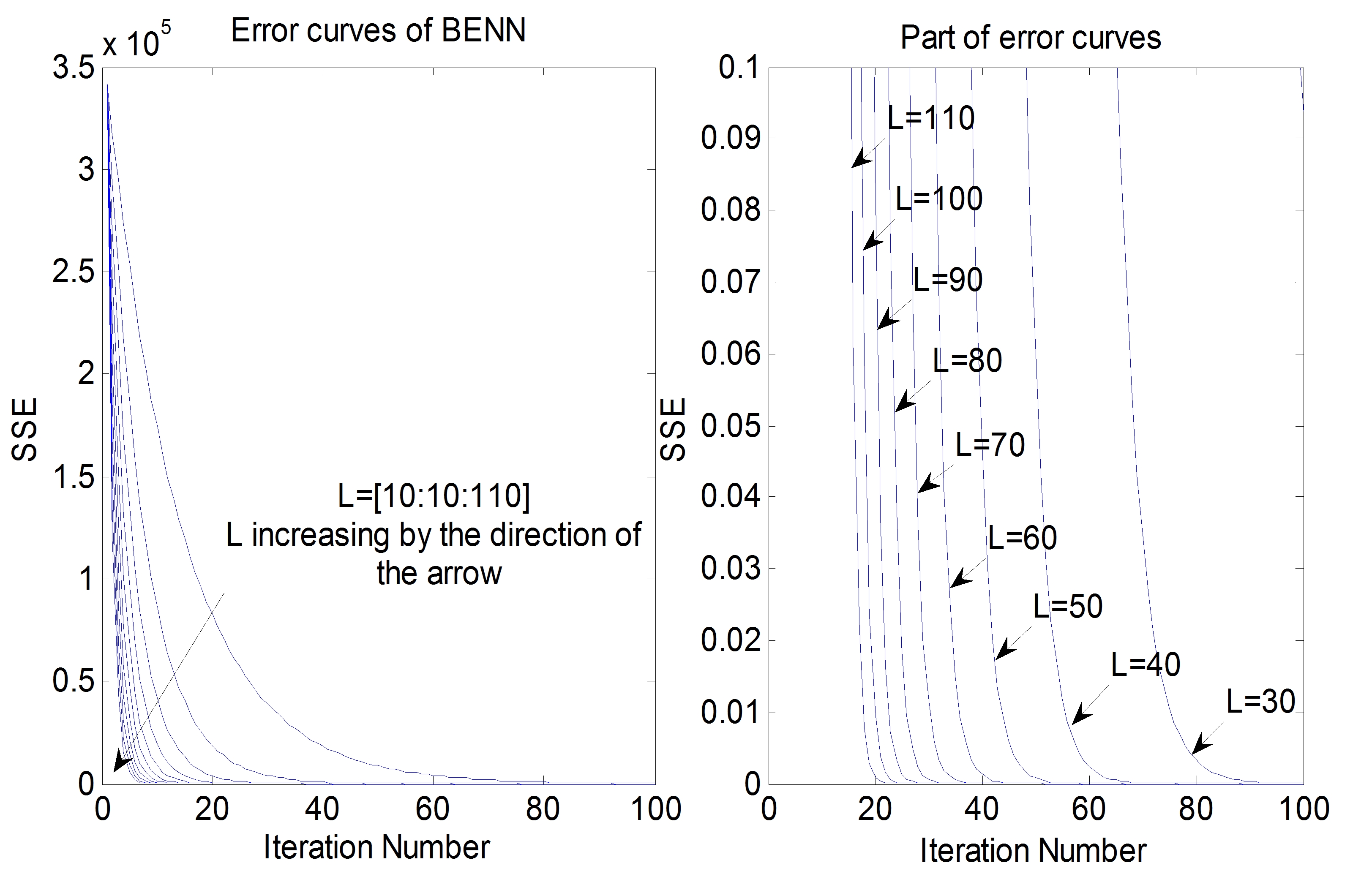}
\caption{Error curves of the BENN model with the increase of $L,L=[10:10:110]$}
\label{5}
\end{figure}

For the EWNN model, make the number of hidden neurons $L$ increases from $10$ to $60$ with the interval of $5$. The error curves of SSE are showed in Fig. 6.
\begin{figure}[h]
\centering
\includegraphics[width=9cm, height=6cm]{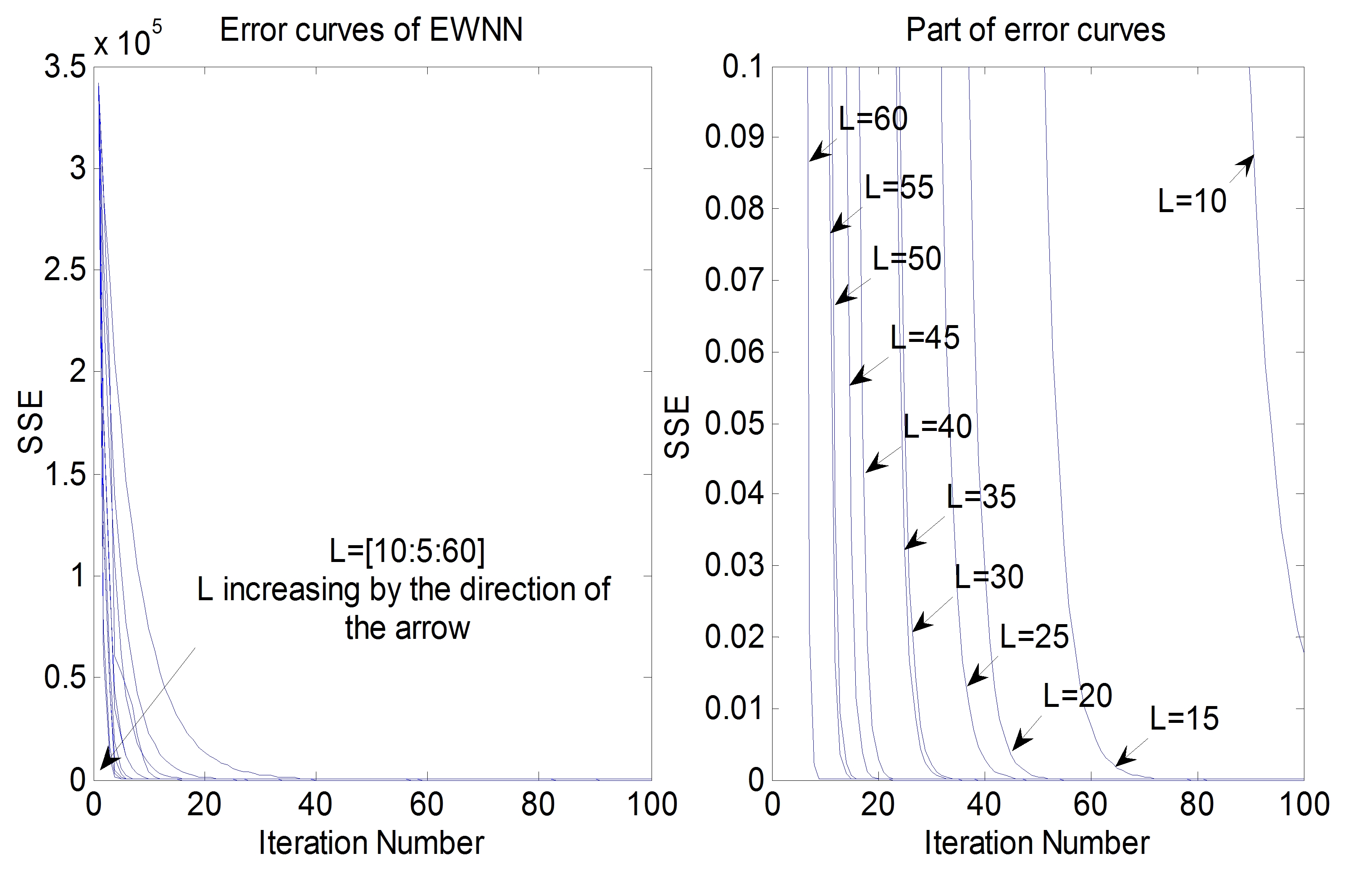}
\caption{Error curves of the EWNN model with the increase of $L,L=[10:5:60]$}
\label{6}
\end{figure}
As showed in Fig. 5 and Fig. 6, with the increase of the iteration number, the error curves of SSE drop rapidly. The larger the number of hidden neurons $L$ is, the faster the error curve drops, and the less iteration number needed to reach the same SSE. The comparison of the part of error curves shows that EWNN model has faster convergence speed than BENN model. When $L=30$, to achieve the SSE of $0.1$, BENN needs about $65$ iteration steps, while EWNN needs only about $25$ iteration steps, which is a big reduction of calculation. To reach the same SSE with the same iteration times, EWNN needs fewer hidden neurons than BENN. For instance, when SSE=0.1 and the iteration number is $30$, BENN needs more than $60$ hidden neurons while EWNN needs only about $25$ hidden neurons.

In the following paper, $L=30$ is selected as the number of hidden neurons which is appropriate for two neural network models according to the error curves.

\subsection{The influence of the update of parameters $a$ and $b$ in EWNN}
When $L=30$, $\epsilon_{min}=10^{-3}$, $\eta_{4}=\eta_{5}=0.01$, a comparison between the EWNN model's error curve with or without the update of $a$ and $b$ of the Morlet wavelets in the hidden layer is showed in Fig. 7. Here $a$ and $b$ are randomly subject to the standard normal distribution.
\begin{figure}
\centering
\includegraphics[width=8cm, height=7cm]{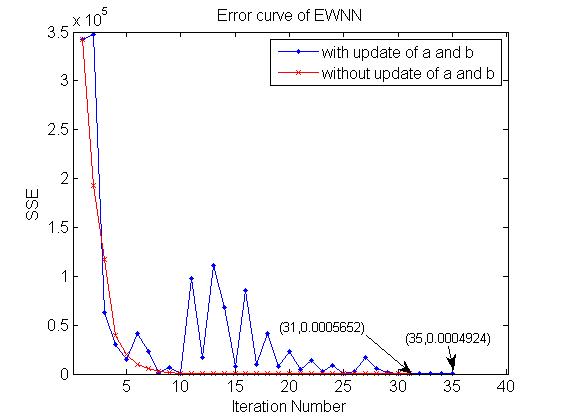}
\caption{Error curve of EWNN when $L=30$, $\epsilon_{min}=10^{-3}$}
\label{7}
\end{figure}
It can be seen in Fig.7 that in the same initial conditions and to achieve the same error threshold, there are a lot of fluctuations in the error curve with the update of $a$, $b$, and inversely increases the number of iteration steps.

The fluctuations have no regular pattern. With different initial value of $a$ and $b$, the performance of SSE changes: Sometimes the SSE has local minimum and leads the training process to a wrong ending or need more iteration steps; In some other cases, it may need fewer iteration steps than the EWNN without the update of $a$ and $b$. But no matter how to set the initial value of $a$ and $b$, the fluctuations are always there. Making comprehensive consideration for the performance of the model, it is better to ignore the update of $a$ and $b$. In the following paper, the value of $a$ and $b$ in EWNN are the same as used in the simulation of Fig.7.
\subsection{Simulation results of three behavioral models}
The Volterra-Laguerre model is proposed in [9] for RF power amplifiers which is also suitable for modeling CDPAs. There are two parameters in this model: the number of Laguerre orthogonal functions $K$ and the pole of Laguerre functions $\lambda,|\lambda|<1$. When $K=3$, this model can't reconstruct the output no matter how to set the value of $\lambda$. In this paper, we choose $K=5$ and $\lambda=0.994$. There are $605$ parameters needed to be estimated.

For the BENN and EWNN model, set the iteration times $N_{max}=40$. Comparisons among the simulation results of the Volterra-Laguerre model, the BENN model and the EWNN model in time domain are given in Fig. 8, the spectrum and spectrum error are showed in Fig. 9.

\begin{figure}
    \centering
    \subfigure[Volterra-Laguerre model]
    {
        \includegraphics[width=8cm, height=6cm]{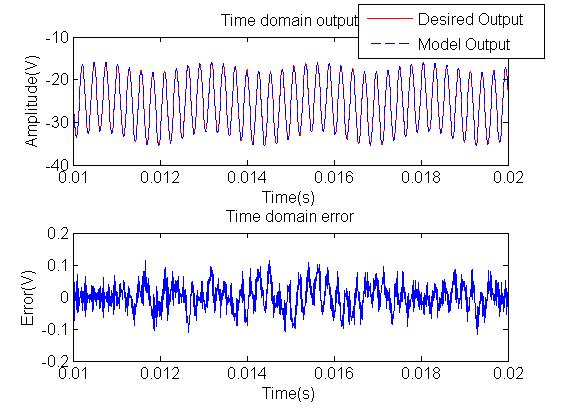}
        \label{fig8(a)}
    }
    \\
    \subfigure[BENN model]
    {
        \includegraphics[width=8cm, height=6cm]{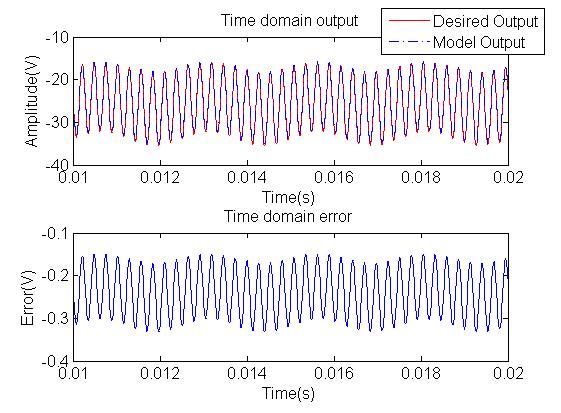}
        \label{fig8(b)}
    }
    \\
    \subfigure[EWNN model]
    {
        \includegraphics[width=8cm, height=6cm]{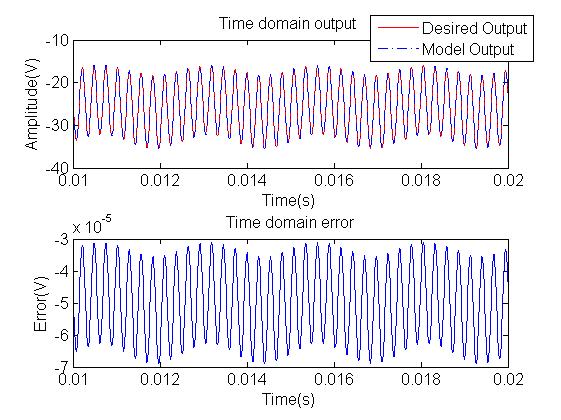}
        \label{fig8(c)}
    }
    \caption{Comparison among three behavioral models in time domain}
    \label{fig8}
\end{figure}

The SSE and the maximum error in time domain of three models are listed in Table 1.
\begin{table}[htbp]
	\caption{SSE and the maximum error of three models in time domain}
	\label{table_five}
	\begin{center}
	\begin{tabular}{|c|c|c|c|} \hline
	Parameter & Volterra-Laguerre & BENN & EWNN  \\ \hline
	SSE & 0.9029 & 37.86 & $2.495 \times 10^{-6}$ \\ \hline
    Max. error(V) & 0.1172 & 0.3311 & $6.8812 \times 10^{-5}$ \\ \hline
		\end{tabular}
	\end{center}
\end{table}

It can be seen in Fig. 8 that the models can reconstructed the output data from the input sinusoidal signal with different accuracy. Comparing EWNN to the Volterra-Laguerre model, there are $605$ coefficients needed to be estimated in the Volterra-Laguerre model and the maximum error in time domain is $0.1172V$ while EWNN only has $30$ hidden neurons, and the time domain error is very subtle. EWNN model has a large reduction in the amount of calculation and much more precise in the model's output than the Volterra-Laguerre model. EWNN is also much more precise than BENN with the same amount of calculation.

\begin{figure}
    \centering
    \subfigure[Volterra-Laguerre model]
    {
        \includegraphics[width=8cm, height=6cm]{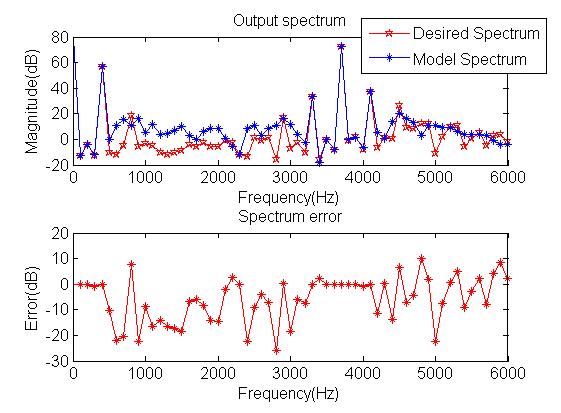}
        \label{fig9(a)}
    }
    \\
    \subfigure[BENN model]
    {
        \includegraphics[width=8cm, height=6cm]{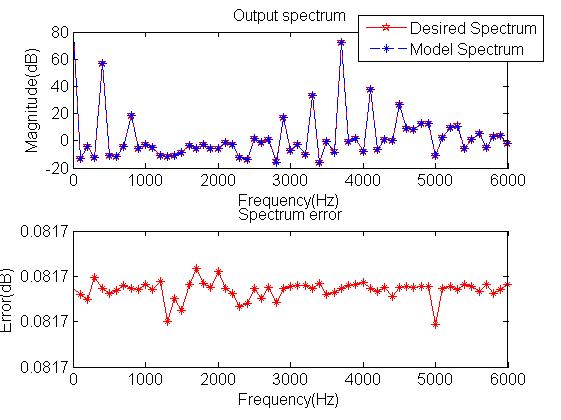}
        \label{fig9(b)}
    }
    \\
    \subfigure[EWNN model]
    {
        \includegraphics[width=8cm, height=6cm]{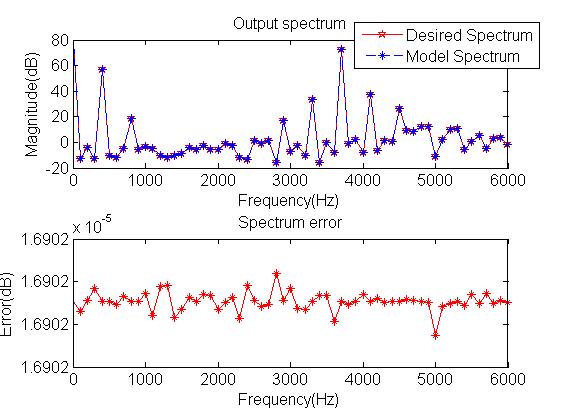}
        \label{fig9(c)}
    }
    \caption{Comparison of three behavioral models in frequency domain}
    \label{fig9}
\end{figure}

The spectrum errors of three models are listed in Table 2. The frequencies are the same as marked in Fig. 2.
\begin{table}[htbp]
	\caption{Measured spectrum and the spectrum error of three models}
	\label{table_five}
	\begin{center}
{\tabcolsep 0.9pt\tiny
	\begin{tabular}{|c|c|c|c|c|c|c|c|} \hline
	$Spectrum(dB)$ &$f_{1}$ &$f_{2}$ &$f_{3}$ &$f_{4}$ &$f_{5}$ &$f_{6}$ &$f_{7}$  \\ \hline
	Measured spectrum & $56.78$ & $18.61$ & $16.98$ & $33.51$ & $72.47$ & $37.54$ & $26.55$ \\ \hline
     Volterra spectrum error & $0.0025$ & $7.532$ & $0.2936$& $0.0177$ & $1.3\times 10^{-5}$ & 0.0430& 6.496 \\ \hline
      BENN spectrum error & $0.0817$& $0.0817$& $0.0817$& $0.0817$& $0.0817$& $0.0817$&  $0.0817$ \\ \hline
    EWNN spectrum error & $1.69\times 10^{-5}$& $1.69\times 10^{-5}$& $1.69\times 10^{-5}$& $1.69\times 10^{-5}$& $1.69\times 10^{-5}$& $1.69\times 10^{-5}$& $1.69\times 10^{-5}$\\ \hline
	\end{tabular}}
	\end{center}
\end{table}

It can be seen that the spectrum errors of the Volterra-Laguerre model fluctuate a lot and are especially large at $f_{2}$ and $f_{7}$, this model loses the correct information of power ripple harmonics and the PS-IMDs in frequency domain. The output spectrum of the basic Elman neural network is $0.08171dB$ smaller than the desired output spectrum at each frequency point. The spectrum of EWNN is the most precise and there is almost no spectrum error.

Fix the number of hidden neurons $L=30$ and set the threshold of SSE $\epsilon_{min}=10^{-3}$, training the basic Elman model and EWNN model, the iteration number needs to reach the SSE threshold and the simulation results are given in Table 3.
\begin{table}[htbp]
	\caption{The convergence of the BENN and EWNN model}
	\label{table_five}
	\begin{center}
	{\tabcolsep 0.9pt
		\begin{tabular}{|c|c|c|c|c|c|c|c|} \hline
	Simulation & Iteration & SSE & Max. time domain & Spectrum error\\
    results & number &  & error(V) & (dB) \\ \hline
	BENN& 86 & 0.000817& 0.0015& 0.000378  \\ \hline
	EWNN& 31& 0.000914& 0.0013& 0.000323 \\ \hline
    	\end{tabular}}
	\end{center}
\end{table}
When the parameters are consistent in these two network models, EWNN has faster convergence speed and less amount of calculation than the BENN. As showed in Table 3, the BENN model needs $86$ iteration steps while EWNN needs only $31$ steps to achieve the same magnitude accuracy, the calculation has been reduced by nearly $64\%$.

\section{CONCLUSION}
In this paper, the asymmetric PS-IMD caused by memory effects is demonstrated and a new behavioral modeling based on the Elman neural network and the wavelets-EWNN is proposed for CDPAs. The Elman network structure manifests the memory effects and the wavelets provide the model fast convergence performance. In the proposed model, the Morlet wavelet functions are employed as the activation function and there is a normalized operation in the hidden layer before the wavelet transform. To obtain more stable convergence performance in SSE curve, the update of the scale factor and translation factor in the wavelet functions are also ignored.
The merits of the proposed model are validated through a comparison with the Volterra-Laguerre model and the BENN model. The simulations carried out in the time and frequency domain indicate that the EWNN model is prior to the Volterra-Laguerre model in both accuracy and calculation. With the same network size, the EWNN model can more accurately characterize PAs than the BENN model and to achieve the same SSE, it's superior to the BENN model in learning speed. Based on the above discuss and the comparison of computer simulation results, the conclusion can be draw that the proposed model is more suitable for analyze both the time domain and the frequency domain nonlinear distortion in PA systems, such as the asymmetric IMD phenomenon. The EWNN model is also appropriate for the RF power amplifiers.
Acknowledgments 

\section{Acknowledgments}
This work was supported in part by the Foundation of Key Laboratory of China's Education Ministry and A Project Funded by the Priority Academic Program Development of Jiangsu Higher Education Institutions.





%

\end{document}